\pdfoutput=1

\documentclass[11pt]{article}

\usepackage[preprint]{coling}

\usepackage{times}
\usepackage{latexsym}
\usepackage{booktabs}

\usepackage[T1]{fontenc}
\usepackage[utf8]{inputenc}
\usepackage{CJKutf8}
\usepackage{microtype}
\usepackage{enumitem}
\usepackage{amssymb}
\usepackage{caption}
\usepackage{graphicx}
\usepackage{url}
\usepackage{longtable}
\usepackage{ascmac}
\usepackage{pgfplots}
\pgfplotsset{compat=1.18}
\usepackage{ulem}
\usepackage{tcolorbox}
\usepackage{pifont}
\usepackage{tabularx}
\usepackage{array}

\newcommand*{\ja}[1]{\begin{CJK}{UTF8}{ipxm}#1\end{CJK}}

\title{IRR: Image Review Ranking Framework for \\ Evaluating Vision-Language Models}

\author{
Kazuki Hayashi\textsuperscript{\dag}, Kazuma Onishi\textsuperscript{\ddag}, Toma Suzuki\textsuperscript{\dag}, \\ \bf
Yusuke Ide\textsuperscript{\dag}, Seiji Gobara\textsuperscript{\dag}, Shigeki Saito\textsuperscript{\ddag}, \\ \bf
Yusuke Sakai\textsuperscript{\dag}, Hidetaka Kamigaito\textsuperscript{\dag}, Katsuhiko Hayashi\textsuperscript{\S}, Taro Watanabe\textsuperscript{\dag} \\
  \textsuperscript{\dag}Nara Institute of Science and Technology \\
  \textsuperscript{\ddag}Hokkaido University \hspace{3pt} \textsuperscript{\S}The University of Tokyo \\
  \texttt{kazuki.hayashi.hl4@naist.ac.jp} \\
  \texttt{\{sakai.yusuke.sr9, kamigaito.h, taro\}@is.naist.jp} \\
  \texttt{katsuhiko-hayashi@g.ecc.u-tokyo.ac.jp}}

\begin{document}
\maketitle

\begin{abstract}
\label{sec：abst}
Large-scale Vision-Language Models (LVLMs) process both images and text, excelling in multimodal tasks such as image captioning and description generation. 
However, while these models excel at generating factual content, their ability to generate and evaluate texts reflecting perspectives on the same image, depending on the context, has not been sufficiently explored.
To address this, we propose \textsc{IRR: Image Review Rank}, a novel evaluation framework designed to assess critic review texts from multiple perspectives. \textsc{IRR} evaluates LVLMs by measuring how closely their judgments align with human interpretations.
We validate it using a dataset of images from 15 categories, each with five critic review texts and annotated rankings in both English and Japanese, totaling over 2,000 data instances.\footnote{The datasets are available at \url{https://hf.co/datasets/naist-nlp/Wiki-ImageReview1.0}}
Our results indicate that, although LVLMs exhibited consistent performance across languages, their correlation with human annotations was insufficient, highlighting the need for further advancements.
These findings highlight the limitations of current evaluation methods and the need for approaches that better capture human reasoning in Vision \& Language tasks.
\end{abstract}

\section{Introduction}
\label{sec：Introduction}
Large language models (LLMs) ~\cite{touvron2023llama2openfoundation, OpenAI_GPT4_2023, vicuna2023} have achieved significant success in NLP tasks. Recently, leveraging these developments, several large-scale vision language models (LVLMs) have been proposed \cite{liu2023visual, liu2024llavanext, ye2023mplugowl, mplug2, bai2023qwenvlversatilevisionlanguagemodel}, demonstrating strong abilities in visual information processing. 
One such application of LVLMs involves generating textual representations of image content, which can fulfill various practical purposes. While these models are capable of producing rich, detailed linguistic depictions of visual scenes, such representations are not constrained to a single “correct” interpretation.

\begin{figure}[]
\centering
\includegraphics[width=0.98\columnwidth]{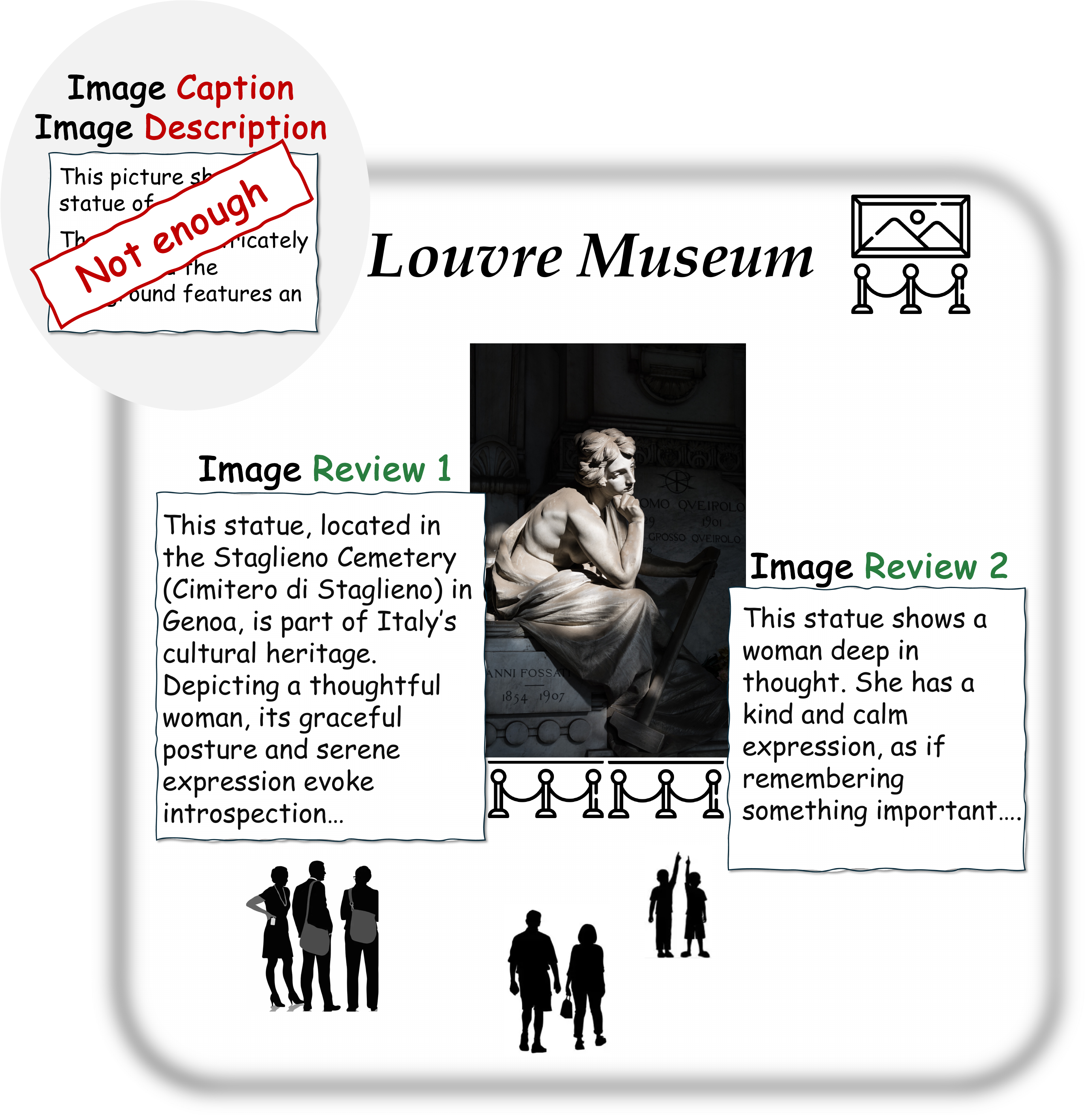}
\caption{Different three image-to-text generation tasks and their corresponding metrics.}
\label{fig:intro}
\end{figure}

In practice, image-related sentences often depend on contextual factors and the viewer’s perspective, with their varying based on according to specific context and focus. 
For instance, as shown in Figure \ref{fig:intro}, adults may emphasize the historical and cultural significance of a statue, while children might prefer to imagine the story behind it or discuss the emotions it evokes. 
These differences highlight the importance of context-specific explanations that can accommodate various perspectives and interests \cite{75effd397a334456a64c85ed56351eff,Naik2023ContextVQATC}.

This inherent diversity in image interpretation introduces significant challenges for evaluating LVLMs. Existing evaluation approaches for image captioning \cite{10.1007/978-3-319-10602-1_48, Desai2020VirTexLV, 10.1007/978-3-030-58577-8_8}, image description \cite{kreiss-etal-2022-concadia, Stangl2021GoingBO}, and image explanation \cite{hayashi-etal-2024-towards,ozaki2024crosslingualexplanationartworklargescale} primarily rely on comparing generated texts with reference texts, assuming a single “correct” answer based on factual elements of the image \cite{7299087, lin-2004-rouge,papineni-etal-2002-bleu}. Such reference-based evaluations are unsuitable for assessing review texts, as they cannot comprehensively capture all valid interpretations, even with multiple references.
Recent studies have explored methods to align model evaluations with human judgments. For instance, pairwise comparisons have been proposed to better reflect human preferences. However, these approaches often fail to account for diverse perspectives. Similarly, reference-free evaluation frameworks have been introduced \citep{pmlr-v235-chen24h, jiang2024genaiarenaopenevaluation, xiong2024llavacritic}. Nevertheless, further validation is needed to determine whether these models \cite{Radford2021LearningTV, ye2023mplugowl, mplug2, liu2024llavanext, bai2023qwenvlversatilevisionlanguagemodel} can effectively capture context and evaluate texts from a human perspective.

To address this, we propose \textsc{IRR: Image Review Rank}, a new evaluation framework designed to evaluate critic review texts from multiple perspectives. 
\textsc{IRR} recognizes various interpretations and evaluates the understanding of LVLMs based on how closely their judgments align with those of humans.
Our method has the model rank five review texts for a given image, comparing the rankings with human rankings to measure correlation. 
This approach assesses whether the model can go beyond factual recognition to identify the most contextually appropriate review.
We constructed a dataset using images from Wikipedia, covering 15 categories.
Each image has five review texts generated by GPT-4V~\cite{OpenAI2023GPT4VisionSC}, manually annotated in English and Japanese, totaling over 2,000 data instances.
Both the English and Japanese datasets were created using the same methodology.

Our results indicate that, although LVLMs performed consistently across languages, their correlation with human annotations reveals room for improvement. Additionally, our framework demonstrates that methods like CLIP \cite{Radford2021LearningTV} are insufficient for evaluating texts in the context of image review. By integrating LLM inferential capabilities with visual information, we better align with human reasoning, further highlighting the limitations of relying solely on CLIP Score \citep{hessel-etal-2021-clipscore}.

\begin{figure*}[ht]
\centering
\includegraphics[width=0.98\textwidth]{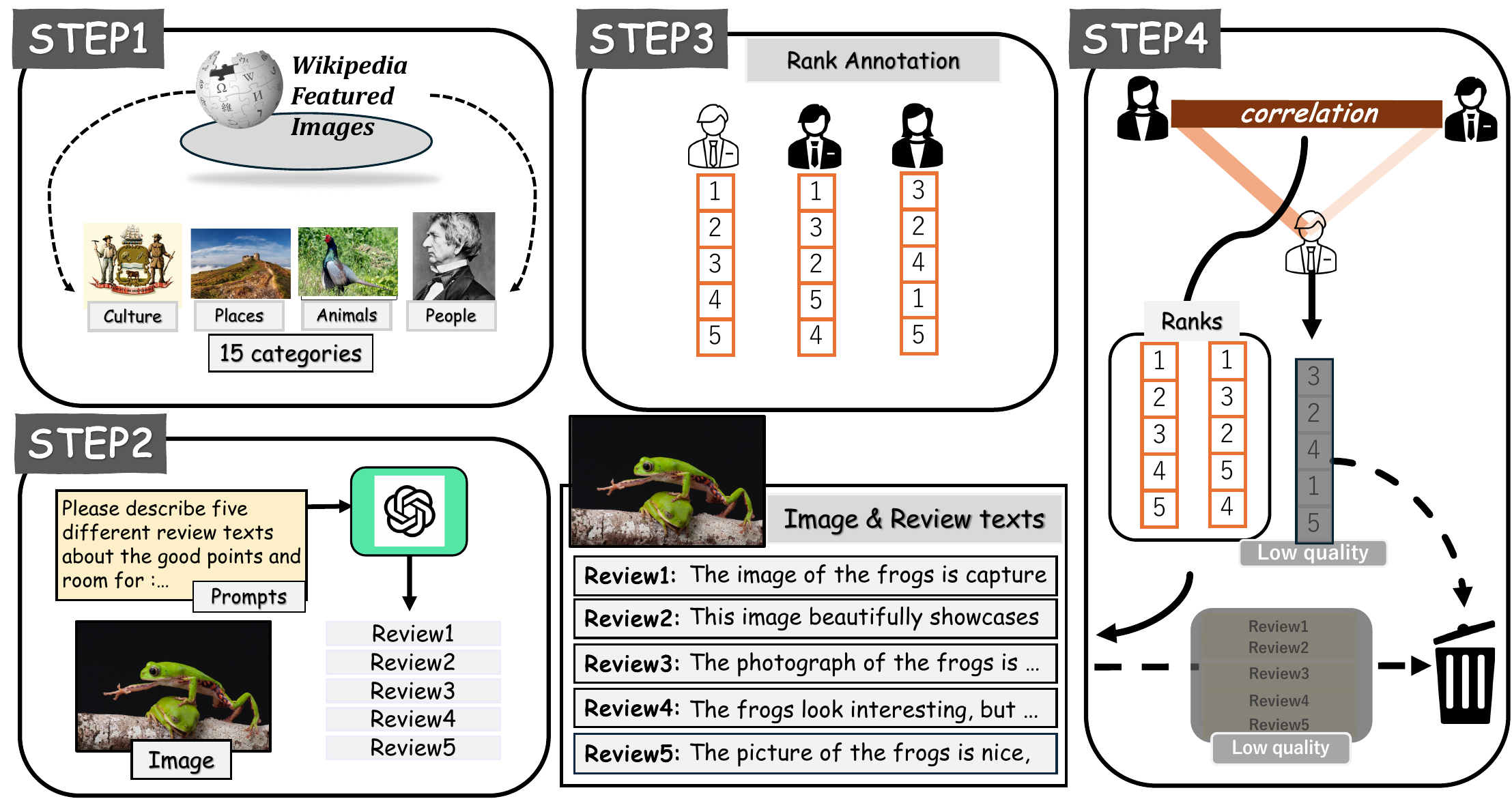}
\caption{Dataset Construction Process.}
\label{fig:Dataset Construction}
\end{figure*}

\section{Evaluation framework}
\label{sec：Image Review Evaluation}

\label{subsec：Evaluation Method}
\subsection{Task: Ranking Review Texts}
\label{paragraph：Perplexity-based Ranking}
We design a task where the model ranks pre-generated texts for a given image based on their relevance within the context of that image. Perplexity is used as the evaluation metric, as it correlates with human judgments of sentence quality \citep{lau-etal-2020-furiously, munoz-sanchez-etal-2024-harnessing}.
Lower perplexity values indicate the model's ability to predict the next word based on context, making it a reliable metric for assessing how well a text aligns with the image context.
To evaluate relevance, perplexity is computed by providing the LVLM with a prefix instruction (see Fig. \ref{appendix:prompt_perplexity}) indicating it is processing a review for an image.
The image and each review text are then input into the model one at a time.
Among the five review texts, those with lower perplexity values are considered better aligned with the image context. The review texts are ranked in ascending order of perplexity.

\subsection{Metric: Measuring Rank Correlation}
\label{paragraph：Measuring Rank Correlation}
To assess how well the LVLM's rankings align with human judgments, we calculate Spearman's rank correlation coefficient \cite{spearman04}. This coefficient ranges from $-1$ (perfect inverse order) to $1$ (perfect alignment) and is computed between the model's rankings and the human rankings for the five texts associated with each image.
Among the three annotators, we first identify the pair with the highest agreement and then calculate the correlation between the LVLM's rankings and each of these two annotators.
The average of these two correlations represents the degree of alignment between the model’s rankings and human evaluations. A higher value indicates a stronger correlation, reflecting the model’s ability to identify high-quality review texts similarly to humans.

\section{Dataset construction}
\label{subsec：Dataset Construction}

The dataset construction process, shown in Figure \ref{fig:Dataset Construction}, was applied to create both the English and Japanese datasets following the same methodology.

\paragraph{STEP 1: Collecting images}
We collected images from the "Featured pictures" section of English Wikipedia.\footnote{\url{https://en.wikipedia.org/wiki/Wikipedia:Featured_pictures}}
This section comprises high-quality images such as photographs, illustrations, and diagrams selected by user votes, covering diverse genres such as artwork, natural landscapes, historical events, and science. We therefore selected it as the image source.
To achieve a diverse selection, we included images from 15 categories, as outlined in Table \ref{tab:categories}.
 
\begin{table}[ht]
\centering
\renewcommand{\arraystretch}{0.70} 
\setlength{\tabcolsep}{4pt} 
\begin{tabularx}{\linewidth}{>{\raggedright\arraybackslash}X>{\centering\arraybackslash}p{3cm}}
\toprule
\rowcolor{black!10} 
\textbf{Category} & \textbf{Number of Items} \\
\midrule
- Animals & 17 \\
- Artwork & 17 \\
- Culture, entertainment, and lifestyle & 16 \\
- Currency & 15 \\
- Diagrams, drawings, and maps & 15 \\
- Engineering and technology & 17 \\
- Natural phenomena & 15 \\
- People & 14 \\
- Places & 17 \\
- Plants & 16 \\
- Sciences & 15 \\
- Space & 15 \\
- Vehicles & 5 \\
- Other lifeforms & 3 \\
- Other & 10 \\
\bottomrule
\end{tabularx}
\caption{Categories and Number of Items.}
\label{tab:categories}
\end{table}

\paragraph{STEP 2: Generating Review Texts}
We used GPT-4V to generate five distinct review texts for each image, reflecting different levels of "Reasonableness" and "Objectivity."
Given the diverse range of image genres, gathering experts or creating texts using external references would have been extremely time-consuming and impractical. Therefore, we opted to use GPT-4V for text generation.
To ensure that the review texts were of varying quality, we designed a prompt to generate five review texts with different levels of reasonableness (see Appendix \ref{appendix:prompt for generating reviews} for details).

\paragraph{STEP 3: Ranking review texts manually}
The five review texts of each image are manually ranked by $X$ ($\geq3$) annotators. English texts were ranked by native or near-native English speakers, and Japanese texts by native Japanese speakers.
To avoid potential biases, the five review texts were randomized before presentation to the annotators. 
Annotators followed detailed instructions (Appendix~\ref{appendix:anotator_instruction}) that emphasized two key criteria: "Reasonableness" and "Objectivity." To guide annotators in selecting the most appropriate text within the context of the image review, "Reasonableness" was further divided into three subcategories: truthfulness, consistency, and informativeness.

\paragraph{STEP 4: Filtering low-quality data}
\label{subsec:filtering}
During annotation, errors caused by misinterpretation, fatigue, or inattention can reduce data quality. To mitigate this, we measure the rank correlations among annotators and filter the data using the correlation scores of the annotator pair with the highest agreement. 
Specifically, we used the correlation coefficient scores described in Section \ref{paragraph：Measuring Rank Correlation} to filter the data. We set the correlation coefficient threshold at $0.6$, retaining only high-quality data with strong inter-annotator agreement (see Appendix \ref{subsec:threshold}).

\section{Experiments}
\subsection{Setup}
\label{subsec:setup}
We evaluated 12 models spanning 7 types of LVLMs:
mPLUG-Owl \cite{ye2023mplugowl}, mPLUG-Owl2 \cite{mplug2}, InstructBLIP \cite{Dai2023InstructBLIPTG}, LLaVA-1.5 \cite{liu2023improvedllava}, LLaVA-NeXT \cite{liu2024llavanext}, Qwen-VL-Chat \cite{bai2023qwenvlversatilevisionlanguagemodel}, and GPT-4 Vision. Additionally, we evaluated six underlying LLMs as the foundation for these LVLMs:
Llama 2 \cite{touvron2023llama2openfoundation}, Vicuna \cite{vicuna2023}, Mistral \cite{jiang2023mistral}, Yi-34B-Chat \cite{ai2024yi}, Qwen-Chat \cite{bai2023qwentechnicalreport}, and GPT-4.
These models were evaluated to compare their perplexity-based ranking in both English and Japanese. For further details, refer to Appendix \ref{appendix:Experimental setting}.
As GPT-4 and GPT-4V lack perplexity measurement for input tokens, we addressed this by ranking five review texts using the same instructions as human annotations based on context-specific relevance (see Appendix \ref{appendix:prompt_rank}).

\begin{table}[t]
\centering
\footnotesize
\resizebox{0.98\columnwidth}{!}{
\begin{tabular}{@{}lrcc@{}}
\toprule
LVLM & Size & EN & JP \\
\midrule
mPLUG\_Owl & 7B & 0.310 & 0.065\\
mPLUG\_Owl 2 & 7B & 0.365 & 0.369\\
InstructBLIP (Vicuna-7B) & 7B & 0.466 & 0.495 \\
InstructBLIP (Vicuna-13B) & 13B & 0.496 & 0.520 \\
LLaVA-1.5 (Vicuna-7B)& 7B & 0.516 & \textbf{0.595} \\
LLaVA-1.5 (Vicuna-13B) & 13B & 0.529 & 0.591 \\
LLaVA-NeXT (Vicuna-7B)  & 7B & 0.510 & \textbf{0.595} \\
LLaVA-NeXT (Vicuna-13B) & 13B & 0.535 & 0.553 \\
LLaVA-NeXT (Mistral-7B) & 7B & \textbf{0.543} & 0.450 \\
LLaVA-NeXT (Yi-34B) & 34B & 0.471 & 0.347 \\
Qwen-VL-Chat \cite{bai2023qwenvlversatilevisionlanguagemodel} & 7B & 0.432 & 0.487\\
\midrule
GPT-4V (Reference) & - & 0.399 & 0.506\\
CLIP Score (Reference) & - & -0.437 & - \\
Human (Reference) & - & 0.795 & 0.846\\
\bottomrule
\end{tabular}
}
\caption{Correlation comparison of LVLMs in English and Japanese. The bold font indicates the best score.}
\label{tab:lvlm_results}
\end{table}

\subsection{Results}
\label{subsec:results}

\paragraph{LVLMs}
\label{para:LVLMs}
Table \ref{tab:lvlm_results} presents the evaluation results for LVLMs. In English, LLaVA-NeXT's Mistral-7B achieved the highest performance, with all models scoring between 0.3 and 0.6, demonstrating a moderate correlation with human judgment \citep{spearman04}.
In Japanese, all LVLMs based on Vicuna performed better in Japanese than in English, with LLaVA-Next and LLaVA-1.5's Vicuna-7B showing the highest score in Japanese. 
These models suggest that, despite being trained only in English (\S\ref{sec：LVLMs}), they have inherited the multilingual understanding abilities of LLMs and can effectively handle Japanese, even in vision-based reasoning tasks \cite{briakou-etal-2023-searching}."
In Table \ref{tab:lvlm_results}, the 'Human Reference' reveals the correlation among human annotations. Comparing this with the correlation obtained from LVLMs indicates that there is potential for improvement in LVLM performance in the conducted evaluation.

\paragraph{LVLMs vs. LLMs}
\label{para:LLMs and LLMs}
To examine whether the models rely solely on text quality or also utilize image information for ranking, we conducted the same evaluation using the underlying LLMs without image inputs. As shown in Table \ref{tab:llm_results}, in English, LVLMs slightly outperformed their corresponding LLMs, indicating that image information improves alignment with human judgments. 
The weak correlation observed with LLMs suggests that while text quality contributes to the rankings, it has limitations.
Notably, the performance gap between LVLMs and LLMs was larger in Japanese; for example, LLaVA-NeXT (Mistral-7B) achieved a correlation of 0.450 in Japanese, while Mistral-7B alone scored 0.194. 
This suggests that LVLMs may have the potential to utilize image information more effectively in Japanese. These findings indicate that multimodal inputs could potentially enhance model performance, particularly in Japanese.

\begin{table}[t]
\centering
\footnotesize
\resizebox{0.98\columnwidth}{!}{
\begin{tabular}{@{}lrcc@{}}
\toprule
LLM & Size & EN & JP \\
\midrule
Llama 2  & 7B & 0.319 & 0.413 \\
Vicuna-7B  & 7B & 0.362 & \textbf{0.422} \\
Vicuna-13B & 13B & 0.358 & 0.365 \\
Mistral-7B & 7B & 0.342 & 0.194 \\
Yi-34B-Chat & 34B & \textbf{0.405} & 0.132 \\
Qwen-Chat  & 7B & 0.386 & 0.386 \\
\midrule
GPT-4 (Reference)& - & 0.384 & 0.478 \\
CLIP Score (Reference) & - & -0.437 & - \\
\bottomrule
\end{tabular}
}
\caption{Correlation comparison of LLMs in English and Japanese. The notations are the same as Table \ref{tab:lvlm_results}.}
\label{tab:llm_results}
\end{table}

\begin{table}[t]
\centering
\small
\setlength{\tabcolsep}{3pt}
\resizebox{\columnwidth}{!}{%
\begin{tabular}{@{}lcccccc@{}}
\toprule
  Threshold & \textless 0 & 0 & 0.2 & 0.4 & 0.6 & 0.8  \\ \midrule
Human Correlation (EN)  & 0.539 & 0.588 & 0.677 & 0.766 & 0.795 & 0.927  \\
GPT-4V Included (EN)  & 0.344 & 0.354 & 0.390 & 0.387 & 0.399 & 0.464  \\ \midrule
Human Correlation (JP)  & 0.712 & 0.728 & 0.780 & 0.824 & 0.846 & 0.942  \\
GPT-4V Included (JP)      & 0.478 & 0.485 & 0.494 & 0.511 & 0.506 & 0.543 \\
\bottomrule
\end{tabular}
}
\caption{Correlation between Human and Human Including GPT-4 Evaluations.}
\label{tab:human_gpt4_results}
\end{table}

\paragraph{Comparison to CLIP Score}
\label{CLIP Score}
Instead of using perplexity for LVLMs, we employed CLIP Score \cite{hessel-etal-2021-clipscore} to measure the alignment between images and text, and used it for scoring and ranking. However, as shown in the table, the CLIP alignment score resulted in negative correlation values.
Therefore, it is clearly shown in our framework that relying solely on image-text alignment methods like CLIP is insufficient for evaluating the appropriateness of texts in the context of image review. We can only achieve alignment with human reasoning by integrating the inferential capabilities of LLMs with visual information. This empirically confirms the limitations of existing studies that use CLIP Score as an evaluation metric.

\paragraph{Comparison between GPT-4V and Humans}
Table \ref{tab:human_gpt4_results} shows the correlation between human annotations and GPT-4V's rankings when GPT-4V is included alongside human annotators. GPT-4V was evaluated using the same prompt as human annotators for direct comparison. To focus on reliably annotated data, we set a threshold on inter-annotator agreement (see Appendix \ref{appendix:promptrank}), retaining only instances where human annotators exhibited high agreement. While the agreement among human annotators increased with higher thresholds, the correlation between GPT-4V's rankings and human annotations showed only a slight improvement in both English and Japanese, remaining approximately 0.5 even at a threshold of 0.8. These results suggest that although GPT-4V generally captures human perspectives, it diverges in some areas, indicating that its ability to rank reviews in alignment with human judgments is still limited. 

\section{Related Work}
\paragraph{LVLMs}
\label{sec：LVLMs}
LVLMs \cite{pmlr-v202-li23q, liu2024llavanext, bai2023qwenvlversatilevisionlanguagemodel, mplug2} integrate a Vision Encoder \cite{pmlr-v202-li23q}, trained through contrastive learning for visual information processing, with Large Language Models (LLMs) \cite{touvron2023llama2openfoundation, vicuna2023, bai2023qwentechnicalreport, jiang2023mistral}.
This integration requires additional training to effectively combine vision and language capabilities. Consequently, these LVLMs outperform conventional pre-trained models, even those with over ten times more parameters \cite{alayrac2022flamingo, pmlr-v202-driess23a}.
However, CLIP \cite{Radford2021LearningTV}, a prominent approach in this domain, primarily aligns images with concise and factual descriptions through contrastive learning. Furthermore, when integrating visual processing capabilities into LLMs via CLIP to construct LVLMs, this method remains limited to factual alignment and lacks the ability to handle diverse perspectives. As a result, LVLMs may be insufficient for generating and interpreting texts that reflect diverse perspectives or identifying the most contextually appropriate ones.

\paragraph{Existing Evaluation Frameworks}
\label{sec：Evaluation Frameworks}
Recent V\&L evaluation frameworks, like Chatbot Arena \cite{chiang2024chatbot} and WildVision Arena \cite{lu2024wildvisionevaluatingvisionlanguagemodels}, use pairwise comparisons to align more closely model outputs with human preferences. While these frameworks  enhance subjective evaluation, they are primarily designed for text-based dialogue or general multimodal tasks and lack multi-perspective consideration.
LLaVA-Critic~\cite{xiong2024llavacritic}, MM-Vet \cite{pmlr-v235-yu24o, yu2024mmvetv2challengingbenchmark}, and MLLM as a Judge \cite{pmlr-v235-chen24h} also utilize LLMs to evaluate LVLMs across various visual tasks focused on factual accuracy and reasoning. However, these frameworks are inadequate for critic review texts requiring various viewpoints. Additionally, their validation for variety and reliability with LLM evaluators is limited, limiting reliable multi-perspective assessments. In contrast, our Image-Review framework provides thorough validation for models like LLaVA-Critic and MLLM-as-a-Judge.

\section{Conclusion}
In this study, we proposed \textsc{IRR: Image Review Rank}, a novel evaluation framework to assess the ability of LVLMs to rank image texts from multiple perspectives and created a corresponding critic review dataset. Our results showed that while LVLMs demonstrated consistent performance across languages, their correlation with human annotations shows room for improvement, highlighting areas for further improvement. Additionally, our framework reveals that methods like CLIP \cite{Radford2021LearningTV} are not adequate for evaluating texts in the context of image reviews. By integrating the inferential capabilities of LLMs with visual information, we improved alignment with human reasoning, further highlighting the challenges of using CLIP Score alone \cite{hessel-etal-2021-clipscore}.
Furthermore, our evaluation framework enables testing model adaptability across contexts by altering the dataset domain. For instance, museum guides require detailed explanations of history and art, while advertising prioritizes concise, attention-grabbing text. Adapting to such diverse contexts remains a key challenge for future research.

\section{Limitations}

\paragraph{Languages.}
In this study, we focused only on English and Japanese, which allowed us to explore the multilingual capabilities of LVLMs. However, it has not been thoroughly investigated whether these findings apply to other languages. One potential limitation is the difference in token length: English had an average token length of 94.77, while Japanese had 50.20, as measured using a multilingual tokenizer. Although Japanese is often considered more compact, direct comparisons are difficult due to structural differences between the languages. This issue reflects the difficulty of controling output lengths by LLMs \citep{juseondo2024instructcmplengthcontrolsentence}. Additionally, the quality of annotations may have varied, as English annotations were provided by non-native speakers, while Japanese annotations were done by native speakers. These factors make it challenging to directly compare accuracy across languages. Note that machine translation may diversify our dataset into multiple languages, whereas it ignores cultural aspects covered by human annotators \citep{sakai-etal-2024-mcsqa}.  

\paragraph{Number of the images.}
The dataset used in this study is domain-specific to Image Review, and it is unclear whether the results can be generalized to other domains. Additionally, the number of images is relatively small, and only test data is provided. Therefore, evaluation using training data or testing generalization capabilities through comparisons with larger datasets remains insufficient.

\paragraph{Prompt for annotations.}
We generated five review texts using GPT-4V with a single prompt, which raises concerns about potential biases introduced by the model \citep{sakai-etal-2024-toward}. Specifically, GPT-4V is known to exhibit a positivity bias~\cite{bender2021dangers}, where certain environmental elements tend to lead to overly positive descriptions. Such biases may result in an imbalanced dataset that does not adequately reflect the diversity of real-world reviews. Additionally, since all reviews were generated solely by GPT-4V, there is a possibility that fundamental biases inherent to the model are embedded in the dataset. While human ranking and filtering of reviews were done manually and through correlation coefficients, these biases still raise concerns regarding the generalizability and neutrality of the dataset.

\paragraph{Entity information.} When LVLMs cannot understand entities in given image reviews, their quality of image review evaluation decreases. Furthermore, image review evaluation requires the alignment of entity knowledge between LLMs and vision encoders similar to other vision and language tasks \citep{kamigaito-etal-2023-table}. Currently, a decisive approach to this problem does not exist. Vision RAG \citep{faysse2024colpaliefficientdocumentretrieval} requires an image review specific datastore. Multimodal Knowledge Graphs (KGs) \citep{chen2024knowledgegraphsmeetmultimodal} are sparse. KG completion (KGC) based on the pre-trained language model (PLM) \citep{yao2019kgbertbertknowledgegraph} suffers from data leakage \citep{sakai-etal-2024-pre}. Traditional KGC models \citep{10.5555/3104482.3104584} are reliable by theoretical studies like \citet{kamigaito-hayashi-2021-unified,pmlr-v162-kamigaito22a,kamigaito22erratum,feng-etal-2023-model,feng-etal-2024-unified}, while their performance is lower than PLM-based ones. 

\section{Ethics Considerations}
\paragraph{Licenses.}
We used Wikipedia materials in the dataset curation process. While Wikipedia text content is available under fair use and the CC-BY-SA 4.0 license\footnote{\url{https://en.wikipedia.org/wiki/Wikipedia:Copyrights}}, we recognize that images have individual licenses. We verified that the images used, particularly from the "Featured Images" section, are covered by the appropriate licenses, including CC-BY-SA when applicable. For transparency, we provide the URL to the source for each image. Additionally, our dataset includes outputs from GPT-4V under OpenAI's license terms\footnote{\url{https://openai.com/policies/terms-of-use}}, granting us full ownership of the generated content.

\paragraph{Moderations and biases.}
In this study, our dataset was created using images obtained from English Wikipedia. The editors of English Wikipedia remove unnecessarily aggressive content\footnote{\url{https://en.wikipedia.org/wiki/Wikipedia:Offensive_material}}, and we also excluded images involving political issues and other sensitive topics from our dataset. 
However, as acknowledged on its official pages\footnote{\url{https://en.wikipedia.org/wiki/Wikipedia:Neutral_point_of_view\#Bias_in_sources},\url{https://en.wikipedia.org/wiki/Wikipedia:Reliable_sources\#Biased_or_opinionated_sources}}, the present English Wikipedia allows the inclusion of information from sources that may be biased. Consequently, the dataset we developed might also reflect the inherent biases of the English Wikipedia.

\bibliography{reference}
\newpage
\appendix
\onecolumn
\section{Details of Prompts}
\subsection{Prompt for Generating Five Review Texts}
\label{appendix:prompt}
We generated five review texts for a certain image using the following prompt, which was designed to create differences among the five review texts.

\begin{tcolorbox}[title=Prompt for Generating Reviews, boxrule=1pt]
\texttt{You are a perceptive and insightful reviewer. Your task is to write five distinct review texts that discuss the strengths and areas for improvement of the given image, while following the constraints below:} \\

\texttt{Guidelines:} \\
\texttt{1. Each review text should present unique content.} \\
\texttt{2. Ensure that the length of each review is approximately equal.} \\
\texttt{3. Do not use bullet points or lists; maintain a cohesive narrative.} \\
\texttt{4. Write reviews in the following order: "Objective and reasonable," "Subjective but reasonable," "Objective but unreasonable," "Subjective and unreasonable," and "Subjective and containing an error."} \\
\texttt{5. Each review should address both the strengths and potential areas for improvement of the image.} \\
\texttt{6. If no improvements are necessary, explicitly state this within the review.} \\

\texttt{Your reviews will contribute to research purposes only and should reflect careful thought and analysis.}
\label{appendix:prompt for generating reviews}
\end{tcolorbox}

\subsection{Prompt for Measuring Perplexity}
\label{appendix:prompt_perplexity}
We measured perplexity for each data using the following prompt.

\begin{tcolorbox}[title=Prefix Sentence, boxrule=1pt]
\texttt{Please describe a review text about the good points and room for improvement of the image.}
\end{tcolorbox}

\subsection{Prompt for Ranking Review Texts}
\label{appendix:prompt_rank}
We input the following prompt into GPT-4V for response-based ranking. 
The content of this prompt is based on the instruction for human annotators in Appendix~\ref{appendix:anotator_instruction}.
\begin{tcolorbox}[title=GPT-4V Ranking Prompt, boxrule=1pt]
\texttt{Below are the images and their review texts. Please rank the review text of each image from 1 to 5, in order of appropriateness. Please note that the numbers from 1 to 5 are not scores but rankings, and the smaller the number, the more appropriate it is. There should be no ties, and each rank from 1 to 5 should always appear once.} \\

\texttt{Please judge the appropriateness by the following aspects in the following order. That is, first, rank the texts by truthfulness. If there are equally truthful texts, rank them by consistency. Similarly, if they are equal also in consistency, rank them by informativeness; if they are equal also in it, rank them by objectivity; if they are equal also in it, rank them by fluency.} \\

\begin{enumerate}
    \item \texttt{Truthfulness: Is it free of false information?}
    \item \texttt{Consistency: Does it correspond to the image?}
    \item \texttt{Informativeness: Does it describe detailed information or features of the image?}
    \item \texttt{Objectivity: Is it an objective description?}
    \item \texttt{Fluency: Is it grammatically correct?}
\end{enumerate}

\texttt{If the text contains unfamiliar information, you may use a dictionary or search engine. However, please do not use a generative AI such as ChatGPT or image search.} \\
\texttt{Do not include the reason for ranking.} \\
\texttt{Absolutely respond in the following format:} \\

\texttt{text1:2nd place} \\
\texttt{text2:3rd place} \\
\texttt{text3:1st place} \\
\texttt{text4:5th place} \\
\texttt{text5:4th place}
\end{tcolorbox}

\newpage
We input the following prompt into GPT-4 for responsed-base ranking without using an image.
\begin{tcolorbox}[title=GPT-4 Ranking Prompt, boxrule=1pt]
\texttt{Please rank the review text by quality.} \\
\begin{center}
\texttt{text1:review text1} \\
\texttt{text2:review text2} \\
\texttt{text3:review text3} \\
\texttt{text4:review text4} \\
\texttt{text5:review text5}
\end{center}

\texttt{Do not include the reason for ranking.} \\
\texttt{Absolutely respond in the following format:} \\
\begin{center}
\texttt{text1:2nd place} \\
\texttt{text2:3rd place} \\
\texttt{text3:1st place} \\
\texttt{text4:5th place} \\
\texttt{text5:4th place}
\end{center}
\end{tcolorbox}

\newpage
\section{Details of Instruction}
\label{appendix:anotator_instruction}
The annotators ranked the review texts according to the following instructions.
\begin{tcolorbox}[title=Instruction, boxrule=1pt]
\texttt{Below are the images and their review texts. Please rank the review text of each image from 1 to 5, in order of appropriateness. Please note that the numbers from 1 to 5 are not scores but rankings, and the smaller the number, the more appropriate it is. There should be no ties, and each rank from 1 to 5 should always appear once.} \\

\texttt{Please judge the appropriateness by the following aspects in the following order. That is, first, rank the texts by truthfulness. If there are equally truthful texts, rank them by consistency. Similarly, if they are equal also in consistency, rank them by informativeness; if they are equal also in it, rank them by objectivity; if they are equal also in it, rank them by fluency.} \\

\begin{enumerate}
    \item \texttt{Truthfulness: Is it free of false information?}
    \item \texttt{Consistency: Does it correspond to the image?}
    \item \texttt{Informativeness: Does it describe detailed information or features of the image?}
    \item \texttt{Objectivity: Is it an objective description?}
    \item \texttt{Fluency: Is it grammatically correct?}
\end{enumerate}

\texttt{If the text contains unfamiliar information, you may use a dictionary or search engine. However, please do not use a generative AI such as ChatGPT or image search.}
\end{tcolorbox}

\twocolumn
\section{Details of Experimental setting}
\label{appendix:Experimental setting}
\subsection{Reproduction Statements}
In the experiments conducted in Section \ref{subsec:results}, we utilized publicly available models for both LVLM and LLM, including mPLUG-Owl~\cite{ye2023mplugowl}, mPLUG-Owl2~\cite{mplug2}, InstructBLIP~\cite{Dai2023InstructBLIPTG}, LLava1.5~\cite{liu2023improvedllava}, LLava-Next~\cite{liu2024llavanext}, Qwen-VL-Chat~\cite{bai2023qwentechnicalreport}, and GPT-4 API ver. 0.28.0~\cite{OpenAI2023GPT4VisionSC}, using their default hyperparameters. 
Additionally, our dataset and code are available at \url{https://github.com/naist-nlp/Hackathon-2023-Summer}. 
For LLMs, we used models such as Llama2~\cite{touvron2023llama2openfoundation}, Vicuna~\cite{vicuna2023}, Mistral~\cite{jiang2023mistral}, Yi-34B-Chat~\cite{ai2024yi}, Qwen-Chat~\cite{bai2023qwentechnicalreport}, and GPT-4.
To ensure a fair comparison of performance across multiple models, all experiments were conducted on an NVIDIA RTX 6000 Ada GPU, using 16-bit quantization to measure Perplexity. However, due to resource constraints, the LLaVA-NeXT (Yi-34B-Chat) model was loaded and inferred using an NVIDIA A100 80GB PCIe in 16-bit quantization. The same settings were applied to each model for performance comparison purposes.

\subsection{LVLM Details}
\begin{center}
\resizebox{\columnwidth}{!}{%
    \begin{tabular}{@{}lll@{}}
        \toprule
        Model                            & Base Model    & HuggingFace Name/OpenAI API                \\
        \midrule
        mPLUG-Owl                         & LLaMA    & MAGAer13/mplug-owl-llama-7b       \\
        mPLUG-Owl2                       & LLaMA2-7B    & MAGAer13/mplug-owl2-llama2-7b     \\
        InstructBLIP (Vicuna-7B)                          & Vicuna-7B       & Salesforce/instructblip-vicuna-7b                  \\
        InstructBLIP (Vicuna-13B)                         & Vicuna-13B      & Salesforce/instructblip-vicuna-13b             \\
        LLaVA-1.5 
        &
        Vicuna-7B&
        liuhaotian/llava-v1.5-7b
        \\
        LLaVA-1.5                      & Vicuna-13B   & liuhaotian/llava-v1.5-13b    \\
        LLaVA-NeXT (Vicuna-7B)                & Vicuna-7B   & liuhaotian/llava-v1.6-vicuna-7b          \\
        LLaVA-NeXT (Vicuna-13B)           & Vicuna-13B  & liuhaotian/llava-v1.6-vicuna-13b        \\
        LLaVA-NeXT (Mistral)                      & Mistral       & liuhaotian/llava-v1.6-mistral-7b               \\
        LLaVA-NeXT (Yi-34B)                     & Yi-34B      & liuhaotian/llava-v1.6-34b              \\
        Qwen-VL-Chat                    & Qwen      & Qwen/Qwen-VL-Chat             \\
        GPT-4-Vision                   & -    &   gpt-4-1106-vision-preview          \\
        \bottomrule
    \end{tabular}%
}
\end{center}
\subsection{LLM Details}
\begin{center}
\resizebox{\columnwidth}{!}{%
    \begin{tabular}{@{}ll@{}}
        \toprule
        Model                            & HuggingFace Name                \\
        \midrule
        Llama2                          & meta-llama/Llama-2-7b                 \\
        Vicuna-7B                             & lmsys/vicuna-7b-v1.5    \\
        Vicuna-13B                  & lmsys/vicuna-13b-v1.5       \\
        Mistral                     & mistralai/Mistral-7B-Instruct-v0.2             \\
        
        Yi-34B               & 01-ai/Yi-34B
\\    
        Qwen-Chat                      & Qwen/Qwen-7B-Chat
        
        \\    
        GPT-4              &       gpt-4-1106-preview   \\
        \bottomrule
    \end{tabular}%
}
\end{center}
\section{Details of Dataset}
\subsection{Correlation Between Prompt Rank and Human Rank}
\label{appendix:promptrank}
The prompt given to GPT-4V (see Appendix~\ref{appendix:prompt}) instructs it to generate the following five types of review texts;
\begin{itemize}
\item "Objective and reasonable,"
\item "Subjective but reasonable,"
\item "Objective but unreasonable,"
\item "Subjective and unreasonable,"
\item "Subjective and containing an error".
\end{itemize}

This order of instructions is defined as \textbf{prompt rank}.
In ranking, human annotators emphasized being reasonable and objective. 
Consequently, if GPT-4V generates review texts precisely following the prompt, we expect a match between the prompt rank and human rank. 

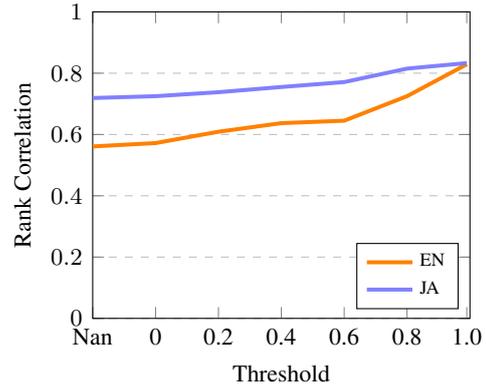
\begin{figure}[t]
    \centering
    \noindent
\pgfplotsset{width= 0.85\linewidth}
\begin{tikzpicture}
\begin{axis}[
    xlabel={Threshold},
    ylabel={Rank Correlation},
    xmin=-0.2, xmax=1,
    ymin=0, ymax=1,
    xtick={-0.2, 0, 0.2, 0.4, 0.6, 0.8, 0.99},
    xticklabels={Nan, 0, 0.2, 0.4, 0.6, 0.8, 1.0},
    ytick={0,0.2,0.4,0.6,0.8,1},
    ymajorgrids=true,
    grid style=dashed,
    xlabel style={font=\small},
    ylabel style={font=\small},
    ticklabel style = {font=\small},
    legend style={at={(0.97,0.03)},anchor=south east,font=\scriptsize},
    legend cell align={left}
]
\addplot[line width=1.5pt,color=orange] coordinates {(-0.2,0.561)(0,0.572)(0.2,0.609)(0.4,0.637)(0.6,0.645)(0.8,0.725)(0.99,0.829)};
\addlegendentry{EN}

\addplot[line width=1.5pt,color=blue!50] coordinates {(-0.2,0.719)(0,0.725)(0.2,0.738)(0.4,0.755)(0.6,0.771)(0.8,0.815)(0.99,0.833)};
\addlegendentry{JA}

\end{axis}
\end{tikzpicture}
    \caption{Correlation between prompt and human ranks.}
    \label{fig:prompt_rank}
\end{figure}

Here, we analyzed the correlation between prompt rank and human rank, and investigated the extent to which GPT-4V can generate review texts following the prompt.
Specifically, we measured the correlation between the prompt rank and top-correlated annotators rank as the threshold was changed.
Figure \ref{fig:prompt_rank} shows the results.

Based on these results, the correlation between prompt rank and human rank showed a strong correlation close to $0.6$ even without setting a threshold.
These findings suggest that there is some validity in the assumption that the 5 review - "objective and consistent," "subjective but consistent," "objective but inconsistent," "subjective and inconsistent," and "subjective and containing errors" - are higher quality in the order of generation in this study's ranking instruction, which emphasizes being reasonable and objective. 
We also found that GPT-4V was able to generate review texts following the prompt to some extent.

\subsection{Review Texts Example}
\label{appendix:reviewexample}
Table \ref{tab:reviewexample}, in English, and Table \ref{tab:reviewexample_ja}, in Japanese, each displays two examples of five review texts generated by GPT-4V.

\subsection{Dataset Construction \& Filtering}
\label{subsec:threshold}
We collected 207 images from 15 different genres, including Animals and Artwork (see Appendix ~\ref{tab:categories} for details).
We generated five review texts for each image in English and Japanese.
The English data were ranked by three native and near-native English speakers, whereas the Japanese data were ranked by three native Japanese speakers.

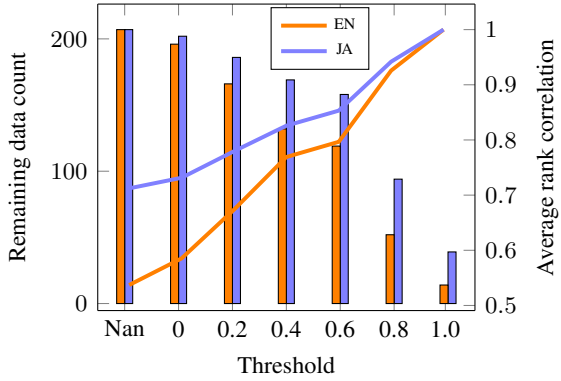
\begin{figure}[t]
    \centering
    \noindent
\pgfplotsset{width= 0.85\linewidth}
\begin{tikzpicture}
\begin{axis}[
    xlabel={Threshold},
    ylabel={Remaining data count},
    xmin=-0.3, xmax=1.1,
    xtick={-0.2, 0, 0.2, 0.4, 0.6, 0.8, 0.99},
    xticklabels={Nan, 0, 0.2, 0.4, 0.6, 0.8, 1.0},
    axis y line*=left,
    ylabel near ticks,
    xlabel style={font=\small},
    ylabel style={font=\small},
    ticklabel style = {font=\small},
    bar width=3pt, 
]
\addplot[ybar, fill=blue!50, bar shift=1.5pt] coordinates {
    (-0.2, 207)
    (0, 202)
    (0.2, 186)
    (0.4, 169)
    (0.6, 158)
    (0.8, 94)
    (1, 39)
};

\addplot[ybar, fill=orange, bar shift=-1.5pt] coordinates {
    (-0.2, 207)
    (0, 196)
    (0.2, 166)
    (0.4, 132)
    (0.6, 119)
    (0.8, 52)
    (1, 14)
};

\end{axis}
\begin{axis}[
    axis y line*=right,
    axis x line=none,
    ylabel={Average rank correlation},
    ytick={0.5, 0.6, 0.7, 0.8, 0.9, 1.0},
    ylabel near ticks,
    yticklabel style={color=black},
    ylabel style={font=\small},
    ticklabel style = {font=\small},
    legend style={at={(0.72,0.99)},anchor=north east,font=\tiny},
]

\addplot[line width=1.5pt,orange] coordinates {
    (-0.2, 0.5370443947)
    (0, 0.5853078924)
    (0.2, 0.6736111111)
    (0.4, 0.7690746753)
    (0.6, 0.7966184807)
    (0.8, 0.9257142857)
    (1, 1)
};
\addlegendentry{EN}

\addplot[line width=1.5pt,blue!50] coordinates {
    (-0.2, 0.7122679072)
    (0, 0.7314653361)
    (0.2, 0.7795576727)
    (0.4, 0.8258448397)
    (0.6, 0.8531405499)
    (0.8, 0.9418820862)
    (1, 1)
};
\addlegendentry{JA}
\end{axis}
\end{tikzpicture}
    \caption{Changes for remaining data count and average rank correlation when varying threshold. The bar graphs represent the remaining data count and the line graphs denote average rank correlation. Nan means no threshold.}
    \label{fig:filtering_result}
\end{figure}
As present in Section \ref{subsec:filtering}, errors may occur in annotations and there is a need for filtering. 
Thus, to identify an optimal threshold for balancing data count and correlation, we varied thresholds over positive values and analyzed changes in data count and average rank correlation between top-correlated annotators (refer to~Section \ref{subsec:filtering}).
The results are shown in Figure \ref{fig:filtering_result}.
Setting the threshold to $0.8$ resulted in a rapid decrease in the remaining data count.
Alternatively, setting the threshold to $0.6$ not only maintained over 100 data count but also showed an average correlation close to $0.8$, a very strong correlation.
Based on these results, we set the optimal threshold to $0.6$.
Furthermore, even when no threshold~(denoted as Nan in Figure \ref{fig:filtering_result}) was set, the average rank correlation was higher than $0.5$. 
The high correlation suggests that annotations within our proposed data construction scheme can be conducted without the need for specialized training of the annotators.

\begin{table*}[htbp]
    \centering
    \small
    \renewcommand{\arraystretch}{1.5} 
    \setlength{\tabcolsep}{5pt} 
    \arrayrulecolor{black!50} 
    \begin{tabular}{c|p{0.75\textwidth}}
    
        \hline
        \rowcolor{black!10} 
        \textbf{Image} & \textbf{Five Review Texts} \\ \hline
        
        \begin{minipage}{30mm}
        \includegraphics[width=1\textwidth]{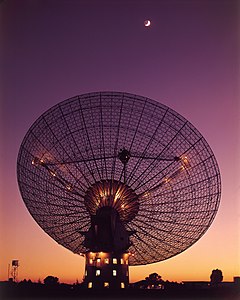}
        \end{minipage} & 
        \begin{minipage}{122mm}
        \vspace{1mm}
        \textbf{1.~Objective and reasonable}: The image captures the exquisite silhouette of a large satellite dish against a vibrant sunset, creating a striking contrast that is visually appealing. The juxtaposition of technology and nature is well executed, showing a blend of human innovation with the beauty of a natural landscape. Room for improvement would include ensuring that the satellite dish is completely in focus to enhance the sharpness of its structure, which is slightly soft around the edges. \\
        
        \textbf{2.~Subjective but reasonable}: This photograph exudes a profound sense of tranquility with the warm hues of the sky complementing the majestic presence of the satellite dish. The composition evokes a sense of wonder and could inspire contemplation about the universe beyond. However, personally, I feel that a wider angle could have included more of the surrounding landscape to further contextualize the dish within its environment. \\
        
        \textbf{3.~Objective but unreasonable}: The technical quality of the image is commendable as it successfully captures the moon in a crescent phase above the dish, which adds an element of scale. Nevertheless, from a strictly compositional standpoint, the satellite appears disproportionately large compared to the moon, which might mislead an uninformed viewer about the actual size differences between the two objects. \\
        
        \textbf{4.~Subjective and unreasonable}: The choice to photograph the satellite dish at dusk is enchanting, as it gives the whole scene a surreal, otherworldly vibe. On the downside, the moon looks disproportionately small, which personally offends my sense of cosmic balance. The photographer should have waited for a night when the moon was bigger or somehow made the satellite dish look smaller to better suit my taste. \\
        
        \textbf{5.~Subjective and containing an error}: The image is enchanting with its blend of purples and blues creating a serene backdrop for the impressive structure of the satellite dish. The way the moon illuminates the scene adds a dreamy quality. However, the image would be more accurate if it showed the dish actively receiving signals, denoted by an unrealistic visual representation like visible beams of light coming down to the dish. \\
        \vspace{1mm}
        \end{minipage} \\ \hline
        
        \begin{minipage}{30mm}
        \includegraphics[width=1\textwidth]{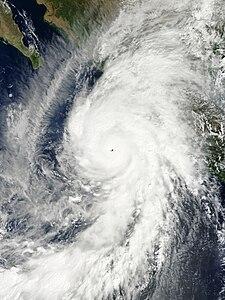}
        \end{minipage} & 
        \begin{minipage}{122mm}
        \vspace{1mm}
        \textbf{1.~Objective and reasonable}: The image effectively captures the formidable structure of a hurricane, showcasing the spiraling clouds and the clear eye at the center, which is indicative of a well-developed storm system. The resolution is high, allowing meteorologists to analyze the storm's features in detail. However, the image is limited in conveying the scale of devastation potential; incorporating overlays indicating size relative to landmasses could enhance its informative value. \\
        
        \textbf{2.~Subjective but reasonable}: From an aesthetic standpoint, the image is breathtaking, with the swirling clouds creating a mesmerizing pattern that is both beautiful and terrifying. The contrast between the storm and the surrounding environment is stark and evocative. On a more subjective note, the image could benefit from a different angle that provides a unique perspective, perhaps highlighting the juxtaposition of the hurricane's might against the vulnerability of the affected areas. \\
        
        \textbf{3.~Objective but unreasonable}: The satellite image is of impeccable quality, presenting the hurricane with precision and clarity, which is crucial for tracking and prediction purposes. However, one might argue that the image should capture more of the surrounding environment to provide a comprehensive view of the storm's reach. This critique, while objective, may be unreasonable considering the limitations of satellite imaging technology and the focus required on the storm itself. \\
        
        \textbf{4.~Subjective and unreasonable}: While the image captures the hurricane's intensity, one might find it lacks dramatic effect. The colors could be more vivid to convey the true ferocity of the storm. This perspective, though subjective, is unreasonable because the purpose of satellite imagery is to provide an accurate representation of reality, not to entertain or exaggerate features for dramatic impact. \\
        
        \textbf{5.~Subjective and containing an error}: The photograph captures what appears to be a tranquil scene, misrepresenting the chaos usually associated with hurricanes. It’s serene and misleadingly calm, which could be improved by adding elements that suggest movement or distress. \\
        \vspace{1mm}
        \end{minipage} \\ \hline
        
    \end{tabular}
    \caption{Examples of five review texts generated by GPT-4V in the English version.} 
    \label{tab:reviewexample}
\end{table*}

\begin{CJK}{UTF8}{min} 
\maketitle
\begin{table*}[htbp]
    \centering
    \small
    \renewcommand{\arraystretch}{1.5} 
    \setlength{\tabcolsep}{6pt} 
    \arrayrulecolor{black!50} 
    \begin{tabular}{c|p{0.75\textwidth}}
    
        \hline
        \rowcolor{black!10} 
        \textbf{Image} & \textbf{Five Review Texts} \\ \hline
        
        \begin{minipage}{30mm}
        \centering
        \includegraphics[width=1\textwidth]{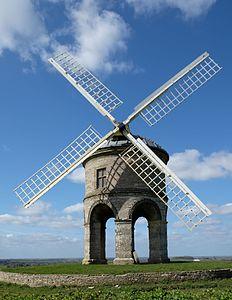}
        \end{minipage} & 
        \begin{minipage}{122mm}
        \vspace{2mm}
        \textbf{1.~Objective and reasonable}: \ja{この画像は明瞭な空の下での風車の構造を鮮明に捉えています。青空を背景にした風車の姿は、建築と自然の調和を表しているようです。また、風車の羽が完全に展開されていることで、その機能性を強調しています。ただし、画像の構図は風車を中心に据えすぎており、周囲の環境にもう少し焦点を当てることで、その場所の広がりや文化的背景に対する理解を深めることができたでしょう。}\\[1mm]
        
        \textbf{2.~Subjective but reasonable}: \ja{この風車の画像を見ると、私は開放感と歴史の重みを同時に感じます。深い青の空が風車の白い羽と石造りの体を引き立てており、そこには一種の詩的な美しさがあります。もし私が撮影者であったならば、風車の足元にもう少し焦点を当て、古い石の質感を表現してみたいと思いますが、この視点もまた、風車の威厳を感じさせるものです。}\\[1mm]
        
        \textbf{3.~Objective but unreasonable}: \ja{この風車の画像は、日中の明るい光の中で撮影されており、細部までくっきりと見えます。しかし、風車のデザインが古典的であるかのように感じる一方で、このような建造物が現代においても依然として機能しているかは疑問です。また、この風車が稼働している場面を捉えることができれば、よりダイナミックな画像になったかもしれません。}\\[1mm]
        
        \textbf{4.~Subjective and unreasonable}: \ja{この画像は、青と緑のコントラストが心地よく、風車の古風な魅力を引き出しています。風車の周囲には花や植物がなく、ある種の孤独感を与えていますが、それがまたこの場所の魔法のように感じます。風車の周囲にもっと生命があれば、画像に暖かみが出て、視覚的にも心にも温かい影響を与えたかもしれません。}\\[1mm]
        
        \textbf{5.~Subjective and containing an error}: \ja{この風車は非常に珍しいデザインで、まるで童話の中から飛び出してきたような不思議な存在感があります。背後に広がる森林が風車の孤立を際立たせています。この写真は、夜に撮影することで神秘的な雰囲気をより高めることができたかもしれませんが、風車の存在そのものがすでに完璧なため、改善の余地はないとも言えます。} \\
        \vspace{2mm}
        \end{minipage} \\ \hline
        
        \begin{minipage}{30mm}
        \centering
        \includegraphics[width=1\textwidth]{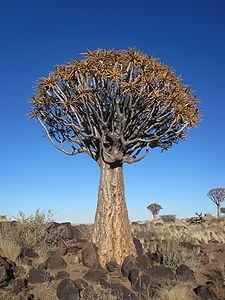}
        \end{minipage} & 
        \begin{minipage}{122mm}
        \vspace{2mm}
        \textbf{1.~Objective and reasonable}: \ja{画像はクリアなブルースカイの背景に、奇妙な形状の樹木を特色としています。樹木の詳細が鮮明にキャッチされており、色彩が鮮やかです。ただし、画像の下部には乱雑な石が多く、これが全体のシンプルさを乱しているように見えます。}\\[1mm]
        
        \textbf{2.~Subjective but reasonable}: \ja{この画像は自然の美しさとユニークな景色を魅力的に捉えています。特に樹木の形状と色が印象的です。しかし、私の感覚では、石の配置が少し散らかって見えるため、それが画像の主要な焦点から視線を逸らしているように感じます。}\\[1mm]
        
        \textbf{3.~Objective but unreasonable}: \ja{画像には大きな樹木があり、背景には青空が広がっています。空は綺麗だが、樹木は他の植物と競合しているように見えます。また、画像の下部は不明瞭で何を示しているのか分かりにくい。}\\[1mm]
        
        \textbf{4.~Subjective and unreasonable}: \ja{私はこの樹木がとても幻想的に見えると感じます。それはまるで別の惑星から来たようです。しかし、背景の青空はちょっと普通過ぎる気がします。そして、地面の石は何か秘密のメッセージを持っているのかもしれません。}\\[1mm]
        
        \textbf{5.~Subjective and containing an error}: \ja{この画像は熱帯のジャングルで撮影されたようです。大きな樹木はジャングルの中心部に位置しており、背景の青空は雨季の始まりを示しているかもしれません。ただ、ジャングルの地面にこんなに多くの石があることは稀ですので、もう少し自然な見た目にすることができるでしょう。} \\
        \vspace{2mm}
        \end{minipage} \\ \hline
        
    \end{tabular}
    
    \caption{Examples of five review texts generated by GPT-4V in the Japanese version.} 
    \label{tab:reviewexample_ja}
\end{table*}
\end{CJK}
\end{document}